\documentclass{article}

 \usepackage[preprint]{neurips_2026}

\usepackage[utf8]{inputenc} 
\usepackage[T1]{fontenc}    
\usepackage{hyperref}       
\usepackage{url}            
\usepackage{booktabs}       
\usepackage{amsfonts}       
\usepackage{amsmath}
\usepackage{pifont}
\usepackage{nicefrac}       
\usepackage{microtype}      
\usepackage{xcolor}         
\usepackage[table]{xcolor}
\usepackage{arydshln}
\usepackage{algorithm}
\usepackage{algpseudocode}
\usepackage{graphicx}
\usepackage{subcaption}
\usepackage{cleveref}
\usepackage{wrapfig}

\title{An Interpretable Latency Model for\\Speculative Decoding in LLM Serving}

\newcommand{\cone}{\ensuremath{C_1}}
\newcommand{\ctwo}{\ensuremath{C_2}}
\newcommand{\conep}{\ensuremath{c_{1,p}}}
\newcommand{\conev}{\ensuremath{c_{1,v}}}
\newcommand{\coned}{\ensuremath{c_{1,d}}}
\newcommand{\ctwop}{\ensuremath{c_{2,p}}}
\newcommand{\ctwov}{\ensuremath{c_{2,v}}}
\newcommand{\ctwod}{\ensuremath{c_{2,d}}}

\newcommand{\acceptrate}{\ensuremath{\alpha}}
\newcommand{\draftlength}{\ensuremath{k}}
\newcommand{\expectedlength}{\ensuremath{E}}
\newcommand{\latency}{\ensuremath{L}}
\newcommand{\rps}{\ensuremath{RPS}}
\newcommand{\batchsize}{\ensuremath{B}}
\newcommand{\decodelength}{\ensuremath{g}}

{
\setlength{\tabcolsep}{0pt}

\author{
\centering
\begin{tabular}{@{}p{0.3\textwidth}@{\hspace{0.03\textwidth}}p{0.3\textwidth}@{\hspace{0.03\textwidth}}p{0.3\textwidth}@{}}
\centering\begin{tabular}[t]{c}
Linghao Kong\thanks{Part of this work was completed during an internship at Red Hat AI.}\\
\normalfont MIT\\
\texttt{linghao@mit.edu}
\end{tabular}
&
\centering\begin{tabular}[t]{c}
Megan Flynn\\
\normalfont Red Hat AI\\
\texttt{mflynn@redhat.com}
\end{tabular}
&
\centering\arraybackslash\begin{tabular}[t]{c}
Michael Peng\\
\normalfont MIT\\
\texttt{mpeng19@mit.edu}
\end{tabular}
\\[3.5em]
\centering\begin{tabular}[t]{c}
Nir Shavit\\
\normalfont MIT,\quad Red Hat AI\\
\texttt{shanir@mit.edu}
\end{tabular}
&
\centering\begin{tabular}[t]{c}
Mark Kurtz\\
\normalfont Red Hat AI\\
\texttt{mkurtz@redhat.com}
\end{tabular}
&
\centering\arraybackslash\begin{tabular}[t]{c}
Alexandre Marques\\
\normalfont Red Hat AI\\
\texttt{almarque@redhat.com}
\end{tabular}
\end{tabular}
}
}

\begin{document}

\maketitle

\begin{abstract}

Speculative decoding (SD) accelerates large language model (LLM) inference by using a smaller draft model to propose multiple tokens that are verified by a larger target model in parallel. While prior work demonstrates substantial speedups in isolated or fixed-batch settings, the behavior of SD in production serving systems remains poorly understood: request load varies over time, and effective batch size emerges from the serving system rather than being directly controlled or observed. In this work, we develop a simple and interpretable latency model for SD in LLM serving. We infer effective batch size from request rate using Little's Law and decompose per-request demand into load-independent and load-dependent components for prefill, drafting, and verification. We validate our model using extensive measurements from vLLM across verifier and drafter model sizes, prefill and decode lengths, request rates, draft lengths, and acceptance probabilities. The model accurately describes observed latency, explains why speedups often diminish as server load increases, and characterizes how draft length, acceptance rate, and verifier-drafter size shape latency across serving conditions, with implications for configuring SD in deployed systems. We further show how the framework extends to mixture of experts models, where sparse expert activation changes the effective service costs across load regimes. Together, our results provide a structured framework for understanding SD in real LLM serving systems.
\end{abstract}

\section{Introduction}

Modern large language models (LLMs) have achieved remarkable performance across a wide range of natural language understanding and generation tasks \citep{grattafiori2024llama, yang2025qwen3}. However, inference costs remain an obstacle for practical deployment. As models scale to trillions of parameters, decoding becomes increasingly expensive: standard autoregressive generation requires sequential token-by-token computation, limiting throughput and increasing latency.

A large body of work aims to accelerate decoding without sacrificing model quality. One prominent approach is speculative decoding (SD), which uses an auxiliary, inexpensive draft model to propose multiple tokens, then validating them in parallel with a larger verifier model \citep{leviathan2023fast, chen2023accelerating}. SD has since been improved via better draft models and verification strategies (e.g. PARD \citep{an2025pard} and EAGLE \citep{li2024eagle1, li2024eagle2, li2025eagle3}) and via complementary directions such as dynamic lookahead \citep{mamou2024dynamic}. In parallel, other acceleration methods seek to generate multiple tokens per forward pass through architectural changes (e.g. Medusa \citep{cai2024medusa}) or via model-free parallel decoding strategies (e.g. Lookahead Decoding \citep{fu2024break}).


Despite this progress, most SD evaluations are performed in simplified single-request or synchronous settings (batch size $=1$) and do not reflect end-to-end behavior in realistic multi-user serving. At scale, SD interacts with system constraints that dominate practical performance. 

\begin{wrapfigure}{r}{0.5\textwidth}
    \centering 
    \includegraphics[width=0.85\linewidth]{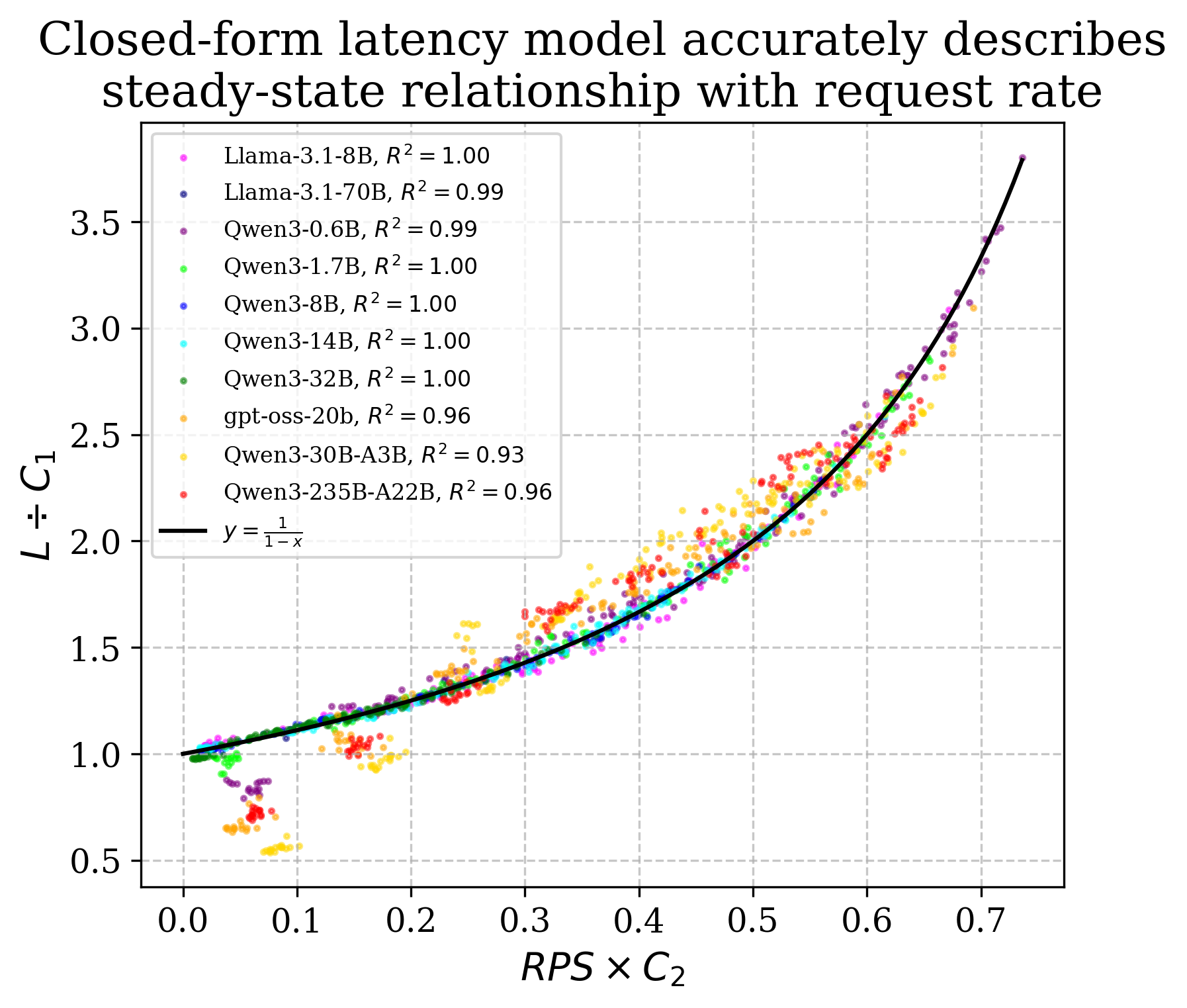}
    \caption{Universal latency scaling across models and configurations. Measured end-to-end request latency \latency{}, normalized by \cone{}, plotted against normalized load $\rps{} \times \ctwo{}$, for multiple models and multiple combinations of prefill/decode configurations. Despite differing absolute scales, all measurements collapse onto the expected curve $y = \frac{1}{1-x}$, validating the latency model (\cref{eq:closed_form_basic}, \cref{sec:steadystate}).}
    \label{fig:basic_latency}
\end{wrapfigure}

For example, serving systems such as vLLM \citep{kwon2023efficient} rely on techniques such as continuous batching and chunked prefill to improve GPU utilization. In such systems, however, batch size is not an experimentally tunable parameter. Rather it emerges dynamically from the request stream and scheduler, changing the critical path that determines end-to-end latency, which can substantially reduce the benefit of speculation or even make SD slower than vanilla decoding. Recent work has started to quantify these effects using metrics such as serving ``goodput'' \citep{liu2024optimizing}, but a simple interpretable model for how SD scales with workload remains lacking. 

In this paper, we study SD under realistic server load by sweeping request rate in an actual vLLM serving stack, where effective batch size is scheduler-dependent rather than directly exposed. We develop scaling trends for SD performance, focusing on how request rate, model size, acceptance rate \acceptrate{}, draft length \draftlength{}, and sequence length jointly shape latency across models. Concretely, our contributions are:

\paragraph{Speculation-aware serving model}
We derive an interpretable latency model for modern, feature-rich LLM servers and extend it to SD via a prefill/verify/draft cost decomposition. This makes explicit how \acceptrate{} and \draftlength{} affect the fixed and load-dependent costs that determine latency under load.

\paragraph{Load-dependent speedup trends}
We derive a closed-form expression for SD speedup as a function of server load, separating the low-load benefit of speculation from the load-dependent terms that dominate near saturation. This explains when speedup improves, erodes, or reverses under realistic serving load, and yields scaling trends across verifier size, drafter size, and sequence length.

\paragraph{Extension across architectures}
We show that the same framework generalizes beyond dense models by introducing a mixture of experts (MoE) aware correction based on expert coverage. This extension explains why sparse models deviate from dense latency scaling at low load and how increasing request rate or \draftlength{} changes their SD behavior.

\section{Related Work}
\paragraph{Speculative decoding} Speculative decoding accelerates autoregressive generation by using a lightweight draft model to propose multiple tokens that are then verified by a larger target model in parallel, reducing the number of expensive target-model decoding steps when draft tokens are accepted \citep{leviathan2023fast, chen2023accelerating}. A growing line of work improves SD by training or adapting draft models and by designing better verification policies to increase acceptance rates and reduce verification overheads \citep{li2024eagle1, li2024eagle2, li2025eagle3, an2025pard, mamou2024dynamic}. In parallel, other approaches pursue multi-token generation per step via architectural changes (e.g., multiple decoding heads) \citep{cai2024medusa} or via algorithmic decoding strategies that weaken sequential dependence without an explicit draft model \citep{fu2024break}. Our work is complementary to these algorithmic advances. Rather than introducing a new SD variant, we model it as a serving-time system component and study how its latency and speedup change under load.

\paragraph{LLM serving systems} Modern serving stacks rely on continuous batching, chunked prefill, and careful KV-cache management to increase GPU utilization and improve throughput. vLLM introduced PagedAttention as an efficient KV-cache management mechanism and is representative of state-of-the-art open serving systems \citep{kwon2023efficient}. More broadly, recent systems work highlights that LLM inference performance depends strongly on how prefill and decode are scheduled, how memory is managed, and how concurrency is controlled under load \citep{agrawal2024taming}. Other work addresses scaling and disaggregation across hardware resources \citep{patel2024splitwise, zhong2024distserve}. SD has also been evaluated from a serving perspective. For example, \citet{liu2024optimizing} optimize SD configurations with a system-level ``goodput'' objective and show that the best SD parameters depend on utilization and workload characteristics. Our analysis is aligned with this systems view, but aims to provide an interpretable descriptive model that explains when and why SD gains diminish or occasionally improve as a function of load, acceptance rate, draft length, and verifier-drafter pairing, without having to explicitly model the granular internals of modern LLM serving stacks.

\paragraph{Queueing-based and systems-aware modeling}
We build on classical queueing insights that relate average concurrency to arrival rate and latency via Little's Law \citep{little1961proof}. In continuous-batching LLM servers, batch size is not directly controlled, but instead emerges from the interaction between request arrivals, scheduler decisions, and per-request service costs. Little's Law provides a way to infer this effective batch size from measurable quantities, yielding a closed-form relationship between request rate and latency. Rather than assuming a particular hardware bottleneck, we model the stable pre-saturation operating regime of the server: from synchronous execution up to the onset of saturation, which we identify by preemption. This provides the foundation for our speculation-aware model, which connects SD algorithmic parameters to load-dependent end-to-end serving behavior.

\section{Methodology}


We use GuideLLM version 0.5.2 \citep{guidellm2024} to drive a vLLM version 0.13.0 \citep{kwon2023efficient} inference server under controlled load and collect end-to-end server-side latency measurements. For each configuration, GuideLLM first measures a synchronous baseline, issuing one request at a time. It then estimates the highest stable request rate before preemption using a throughput benchmark with many concurrent in-flight requests. Finally, it runs a series of eight evenly spaced constant RPS benchmarks between these two endpoints. In practice, we discard the final throughput ceiling, where latency becomes unstable as the server approaches saturation and begins to preempt requests. This produces a systematic sweep of nine total request rates, from synchronous execution to high-utilization, while excluding the unstable saturation boundary.

We sweep prefill and decode lengths over $\{256,512,768,1024\}$ tokens, yielding 16 prefill--decode combinations, using simulated token inputs from Jane Austen's \textit{Pride and Prejudice} \citep{austen1813pride}. For SD, we sweep verifier acceptance rates from $50\%$ to $100\%$ and draft lengths from 1 to 10 tokens, enforcing acceptance rates by overriding vLLM's rejection sampling strategy. We evaluate Llama-3.1-8B-Instruct, Llama-3.1-70B-Instruct \citep{grattafiori2024llama}, gpt-oss-20b \citep{agarwal2025gpt}, and the Qwen3 family \citep{yang2025qwen3}. Except for Llama-3.1-70B-Instruct, SD uses an EAGLE-3-style lightweight drafter \citep{li2025eagle3}. The MoE models gpt-oss-20b, Qwen3-30B-A3B, and Qwen3-235B-A22B are included to evaluate extensions to expert-routed architectures.


To demonstrate the generality of the formulation described here, we also analyzed vanilla SD, where an independent model is used to draft tokens. Specifically, we considered using a Llama-3.1-8B-Instruct drafter for Llama-3.1-70B-Instruct, and a Qwen3-1.7B drafter for Qwen3-14B. Since we focus on the more efficient EAGLE-3 SD, we analyze only 1024 tokens for both the prefill and decode length. At the time of writing, these two types of SD are the most stably supported in vLLM.

All experiments are conducted on a single NVIDIA A100 SXM GPU, except for Llama-3.1-70B-Instruct, which uses 4 NVIDIA A100 SXM GPUs, and Qwen3-235B-A22B, which uses 8 NVIDIA A100 SXM GPUs. To assess hardware robustness, we also evaluate the dense Qwen3 models on a single NVIDIA H100 GPU. We fit all latency models using SciPy \texttt{curve\_fit} \citep{2020SciPy-NMeth}.


\section{Analysis}
\subsection{Steady Server State Formulation of Latency} \label{sec:steadystate}

We first motivate our analysis with a simple latency model for realistic LLM inference servers under load. As stated earlier, in production-style systems such as vLLM, batch size is usually not an explicit control variable. Instead, requests are continuously admitted and scheduled, and the number of concurrently active sequences emerges from the interaction between arrival rate, service latency, KV-cache pressure, and scheduler decisions. Although these mechanisms are complex at the implementation level, we find that their average effect in the stable pre-saturation regime is well captured by a low-dimensional steady-state model.

Our model is inspired by roofline-style cost decompositions \citep{williams2009roofline}, separating latency into load-independent and concurrency-dependent costs. For example, a linear layer incurs fixed costs from preparing weights and batch-dependent costs from applying them to multiple activations. We apply this abstraction to end-to-end request latency: $\latency{} = \cone{} + \batchsize{} \times \ctwo{}$, where \cone{} is load-independent and \ctwo{} is the marginal cost per effective batch element. These coefficients should be interpreted as effective system-level quantities: they absorb the average effects of scheduling, batching, and model execution rather than describing any single kernel or serving mechanism in isolation.

Although \batchsize{} is not directly revealed, Little's Law \citep{little1961proof} relates it to measurable quantities. In a stable system, the long-term average number of active requests equals the request arrival rate multiplied by the average time each request spends in the system: $\batchsize{} = \rps{} \times \latency{}$. Substituting this into the latency model yields a closed-form expression for latency as a function of request rate:
\begin{equation} \label{eq:closed_form_basic}
\latency{} = \cone{} + \rps{} \times \latency{} \times \ctwo{}
\Rightarrow
\latency{} =
\frac{\cone{}}{1 - \rps{} \times \ctwo{}} .
\end{equation}

This formulation is not intended to describe the saturation boundary, where preemption and scheduler effects can cause latency to vary sharply. Instead, we fit and evaluate it from synchronous execution up to the onset of saturation. Within this stable pre-saturation regime, the model accurately captures the average end-to-end request latency behavior of non-speculative serving across model sizes, architectures, and sequence lengths (\cref{fig:basic_latency}). This empirical convergence provides the foundation for our subsequent analysis of SD. MoE models exhibit a systematic deviation at low load, with lower than predicted synchronous latency; we return to and explain this behavior in \cref{sec:moes}.

\subsection{Speculative Decoding Follows Little's Law}

\begin{figure}[H]
    \centering 
    \includegraphics[width=0.25\linewidth]{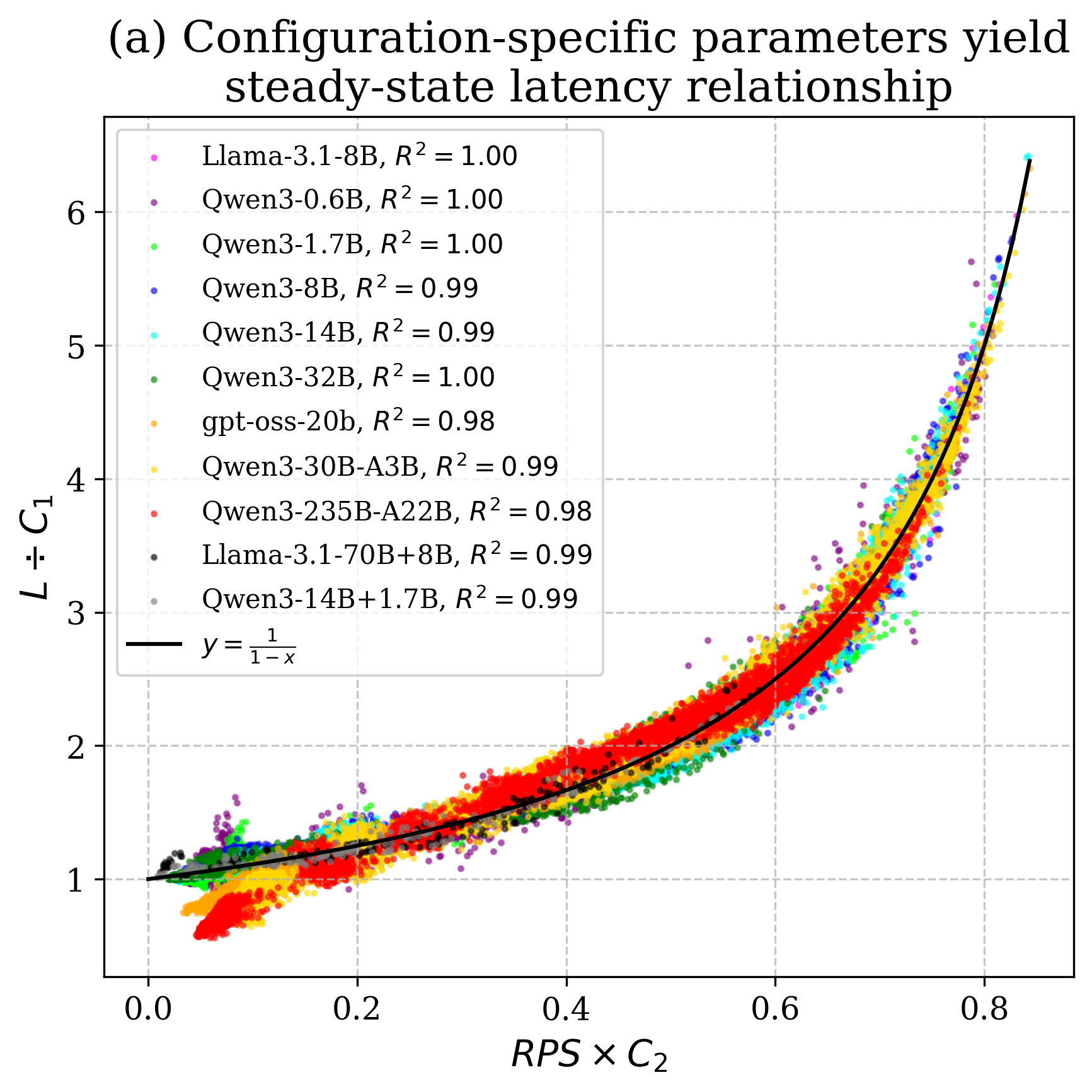}
    \includegraphics[width=0.5\linewidth]{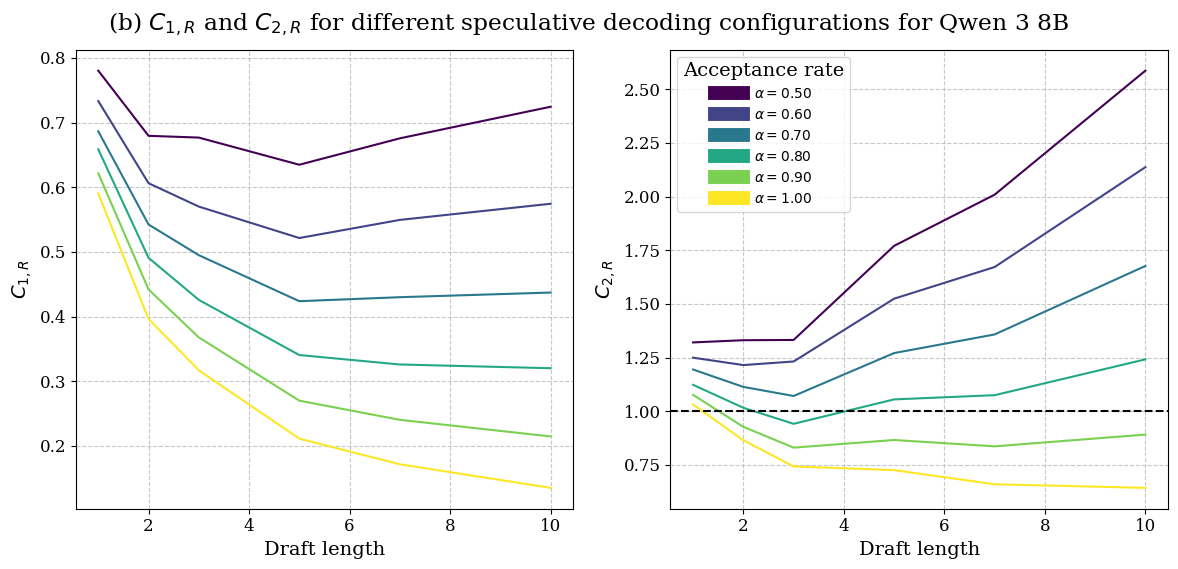}
\caption{
Configuration-dependent latency and speedup behavior under SD.
(a) Per-configuration fits collapse onto the Little's-Law form in \cref{eq:closed_form_basic}, suggesting that SD rescales effective service parameters without changing the underlying load--latency relationship. ``70B+8B'' and ``14B+1.7B'' denote vanilla SD with independent drafters; all other models use EAGLE-3 SD.
(b) For Qwen3-8B at 1024 prefill and decode tokens, $C_{1,R}<1$ across settings, while $C_{2,R}>1$ except at very high acceptance rates, explaining why speedup often erodes under load.
}
    \label{fig:speedup_specific}
\end{figure}

We now fit the steady-state latency formulation from \cref{eq:closed_form_basic} to SD. Here, we fit one separate model per acceptance rate and draft length configuration, in addition to the previous combinations of prefill and draft length. Across all models, prefill–decode combinations, and SD settings, the normalized latencies once again collapse onto the same steady-state curve (with MoEs again showing deviation that is analyzed in \cref{sec:moes}) indicating that SD preserves the underlying serving dynamics  (\cref{fig:speedup_specific}a).

\subsection{Modeling Speedup via Speculative Decoding}

Because both autoregressive models and models augmented with SD are well described by the same closed-form structure, we can express the expected speedup as the ratio of their modeled latencies. Let $\latency{}_D = \frac{C_{1,D}}{1 - \rps{} \times C_{2,D}}$ be the end-to-end request latency of serving a dense model and let $\latency{}_{SD} = \frac{C_{1,SD}}{1 - \rps{} \times C_{2,SD}}$ be the latency of serving a model with SD. The speedup attainable with SD is therefore described by the following equation, with the full derivation in \cref{sec:speedupderivation}:

\begin{equation} \label{eq:speedup}
\begin{split}
\text{Speedup}&=\frac{\latency{}_D}{\latency{}_{SD}}=\frac{1}{C_{1,R}}\times(1 + (1-C_{2,R})\times \frac{r}{1-r})\\
C_{1,R} &= \frac{C_{1,SD}}{C_{1,D}},C_{2,R} = \frac{C_{2,SD}}{C_{2,D}}, r = \rps{} \times C_{2,D}
\end{split}
\end{equation}

This decomposition makes the roles of the parameters explicit. $C_{1,R}$ acts as a multiplicative modulator of speedup, setting the zero-load limit to $1/C_{1,R}$, which reflects the cost of fixed latency components under SD relative to dense decoding. Load dependence enters through the hyperbolic factor $\frac{r}{1-r}$, capturing the rapid growth of latency effects toward saturation. The sign of the load effect is controlled only by $1-C_{2,R}$, as $\frac{r}{1-r}$ is nonnegative. Consequently, speedup increases with load when $C_{2,R}<1$, indicating that SD reduces load-dependent service demand, and decreases with load when $C_{2,R}>1$, matching the common observation that SD benefits erode at high utilization.

Examining the dependence of $C_{1,R}$ and $C_{2,R}$ for a fixed prefill-decode configuration (\cref{fig:speedup_specific}), we find that $C_{1,R}$ is always below 1, consistent with observations that SD performs well in the synchronous regime: several load-independent costs (e.g., weight loading) are amortized across fewer verifier steps. $C_{2,R}$ behavior is more nuanced. For (often unrealistically) high acceptance rates, we observe $C_{2,R}<1$, implying that speedup increases with load. However, for many acceptance-rate and draft-length settings, $C_{2,R}>1$, indicating that speedup diminishes with utilization. Increasing draft length also generally increases $C_{2,R}$, exacerbating this erosion of speedup at high load.

\begin{figure*}[h]
    \centering 
    \includegraphics[width=\linewidth]{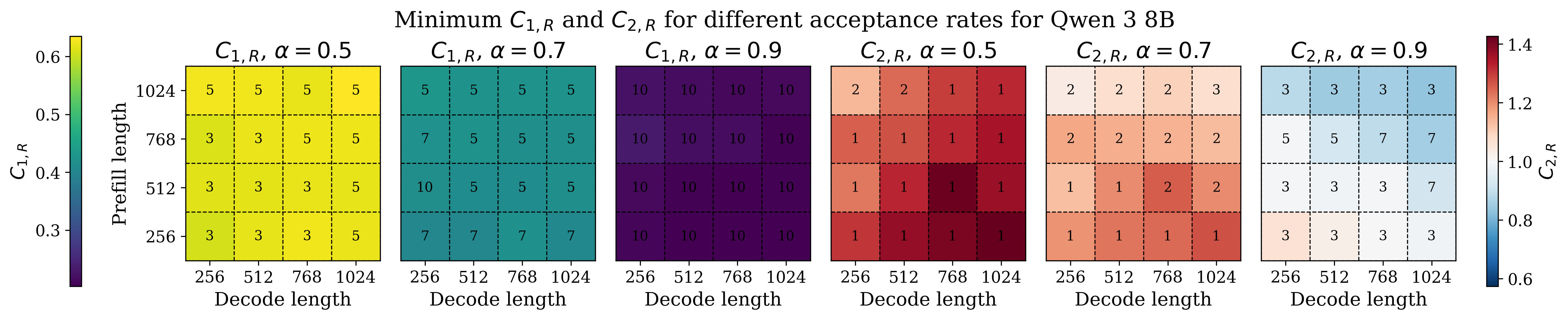}
    \caption{The minimum possible fixed and load-variable cost ratios across prefill and decode configurations. For $\acceptrate{} \in \{50\%, 70\%, 90\%\}$, we show the minimum values of $C_{1,R}$ and $C_{2,R}$, and also inscribe which draft length yielded that minimum value. Other \acceptrate{} shown in \cref{fig:speedup_minimums_full}.}
    \label{fig:speedup_minimums}
\end{figure*}

Next, we consider all prefill-decode settings simultaneously. For each setting and acceptance rate, we compute the minimum values of $C_{1,R}$ and $C_{2,R}$ over all draft lengths. This again shows that $C_{1,R}$ remains consistently below 1, whereas $C_{2,R}$ typically exceeds 1 even in its most favorable configuration for a given acceptance rate. Thus, speedup generally decreases with load, with the exception of the highest-\acceptrate{} regimes, where $C_{2,R}$ can fall below 1 (\cref{fig:speedup_minimums}).

The draft length that minimizes $C_{1,R}$ is also consistently much larger than the draft length that minimizes $C_{2,R}$. This reflects an inherent trade-off in SD. At low load, latency is dominated by the batch-independent term, so larger \draftlength{} can be beneficial by amortizing verifier overhead over more accepted tokens, improving $C_{1,R}$. In contrast, under load the batch-amplified term becomes dominant, and a large \draftlength{} worsens $C_{2,R}$, causing speedup to erode more rapidly with load. Thus, draft lengths optimized for batch size = 1 performance are generally suboptimal for high throughput, emphasizing the need for a model that can describe performance across both load and SD configurations.

\subsection{Latency Model of Speculative Decoding Across Configurations}

\begin{figure*}[h]
    \centering 
    \includegraphics[width=\linewidth]{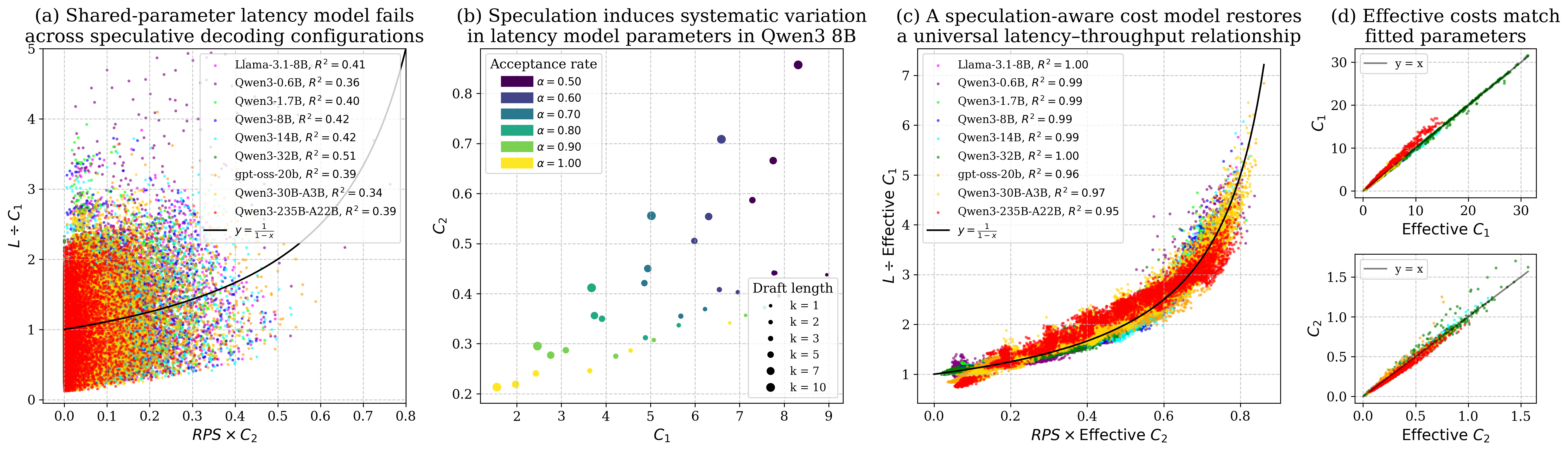}
    \caption{Speculative decoding induces configuration-dependent latency costs which a single speculation-aware cost model effectively captures. (a) Forcing all acceptance rates and draft lengths to share a single $(\cone{}, \ctwo{})$ yields poor fits, indicating that SD alters effective service demand. (b) Fitted parameters for Qwen3-8B vary systematically with acceptance rate \acceptrate{} and draft length \draftlength{}, motivating a speculation-aware cost model. This was modeled at 1024 prefill and 1024 decode tokens. (c) A single speculation-aware cost model describes effective latency parameters across all acceptance rate and draft length combinations using $C_{1,EFF}$ and $C_{2,EFF}$. (d) $C_{1,EFF}$ and $C_{2,EFF}$ closely match the \cone{} and \ctwo{} parameters for a latency model following \cref{eq:closed_form_basic} fitted per \acceptrate{} and \draftlength{} combination.}
    \label{fig:specdec_basic_latency}
    \vspace*{-3mm}
\end{figure*}

To move beyond speedup trends and toward a model of speedup, we first ask whether the same closed-form latency model (\cref{eq:closed_form_basic}) can be fit across all acceptance rates and draft lengths for a fixed system configuration. In this setting, the model fails to capture the observed behavior, suggesting that SD parameters change the effective serving relationship (\cref{fig:specdec_basic_latency}a).

However, as shown before, when fit a separate model for each acceptance rate and draft length, the fit improves substantially, indicating that the underlying model remains valid but with parameters that depend on speculation behavior (\cref{fig:speedup_specific}a). In particular, distinct trends emerge within each configuration: higher \acceptrate{} consistently reduces the effective \cone{} and \ctwo{} costs, whereas the impact of \draftlength{} exhibits a more nuanced dependence (\cref{fig:specdec_basic_latency}b). This observation motivates a refined model that explicitly accounts for how the mechanisms of SD affect per-request service demand.

We begin with a per-cycle view of SD. Conceptually, we split latency into prefill ($c_p$), verification ($c_v$), and drafting ($c_d$) components. After the prefill stage, each decoding cycle consists of two phases: a draft phase, in which the drafter model generates \draftlength{} candidate tokens sequentially, and a verify phase, in which the verifier model evaluates these candidates in parallel. The latency of a single SD cycle can therefore be expressed as $c_v + k \times c_d$, where $c_v$ denotes the verifier latency and $c_d$ the latency of the drafter. For an expected acceptance length of \expectedlength{}, each decode step is amortized over the \expectedlength{} tokens. Therefore, the total number of SD cycles required, on average, is $\frac{\decodelength{}}{\expectedlength{}}$ cycles.


Under this decomposition, the effective load-independent cost can be written as
$
C_{1,\mathrm{EFF}} = \conep + \frac{\decodelength{}}{\expectedlength{}}\left(\conev + k \times \coned\right),
$
where $\conep$, $\conev$, and $\coned$ denote the fixed prefill, verification, and drafting costs, respectively. Similarly, the effective load-dependent cost is
$
C_{2,\mathrm{EFF}} = \ctwop + \frac{\decodelength{}}{\expectedlength{}}\left(\ctwov + k \times \ctwod\right),
$
where $\ctwop$, $\ctwov$, and $\ctwod$ capture the corresponding load-dependent service demands. We model the acceptance process by assuming a constant, per-token acceptance probability \acceptrate{}, independent of previously accepted tokens \citep{leviathan2023fast, li2024eagle1, an2025pard, chen2023accelerating}. The expected number of accepted tokens \expectedlength{} per SD cycle is then $\expectedlength{} = \frac{1 - \acceptrate{}^{\draftlength{}+1}}{1 - \acceptrate{}}$.

Although each verification step evaluates $k$ candidate tokens, our measurements do not show a strong linear dependence of the load-dependent verification coefficient on $k$. Instead, the effect of increasing draft length is better captured by the explicit drafting term $k\ctwod{}$, while the verifier contribution remains approximately constant across draft lengths. We therefore treat $\ctwov{}$ as independent of $k$ in the load-dependent cost model.


Substituting these effective costs into the closed-form latency expression yields:

\begin{equation}
\begin{split}
\label{eq:intermediate}
\latency{} = \frac{\conep{} + \frac{g}{E}(\conev{} + k \times \coned{})}{1 - \rps{} \times (\ctwop{} + \frac{g}{E}(\ctwov{} + k \times \ctwod{}))}
\end{split}
\end{equation}

This formulation makes explicit how SD reshapes the fixed and load-dependent components of latency through $E$. In particular, speculation reduces the effective number of verifier decode steps while introducing additional draft computation, resulting in a tradeoff governed jointly by $k$ and $\expectedlength{}{}$.


When incorporating SD parameters into the end-to-end request latency model, the unified serving relationship is recovered across all acceptance rates and draft lengths: configurations no longer diverge when expressed in terms of the effective fixed and load-dependent costs (\cref{fig:specdec_basic_latency}c). To quantify this agreement, we compare the effective fixed-cost and batch-dependent coefficients $C_{1,EFF}$ and $C_{2,EFF}$ against the values of $\cone{}$ and $\ctwo{}$ obtained by fitting the basic request latency model separately for each configuration. We find close agreement between the two across all settings, indicating that the proposed reparameterization accurately captures how SD reshapes per-request costs (\cref{fig:specdec_basic_latency}d). 

This also suggests a direct deployment use case of the model: because \draftlength{} is a small bounded integer, a serving system can inexpensively estimate the current \acceptrate{} and \rps{}, predict latency for each feasible \draftlength{}, and choose the best value as load or acceptance behavior changes. We further show that the model fits H100 measurements (\cref{sec:h100}) and remains a good approximation for p95 latency and for p99 latency, especially on larger models (\cref{sec:tail_latency}), leaving full distributional modeling to future work.

\subsection{Scaling of Latency Coefficients}
\label{sec:scaling}

We fit \cref{eq:intermediate} to the Qwen3 model family, using a separate latency model for each prefill/decode configuration and pooling measurements across acceptance rates and draft lengths. For Qwen3-8B and Qwen3-14B, we also construct two-, three-, and four-layer drafters in addition to the standard single-layer EAGLE-3 drafter \citep{li2025eagle3}. This lets us study how \conep{}, \conev{}, \coned{}, \ctwop{}, \ctwov{}, and \ctwod{} scale with verifier size, drafter size, and sequence lengths.

\begin{figure*}[h]
    \centering 
    \includegraphics[width=0.8\linewidth]{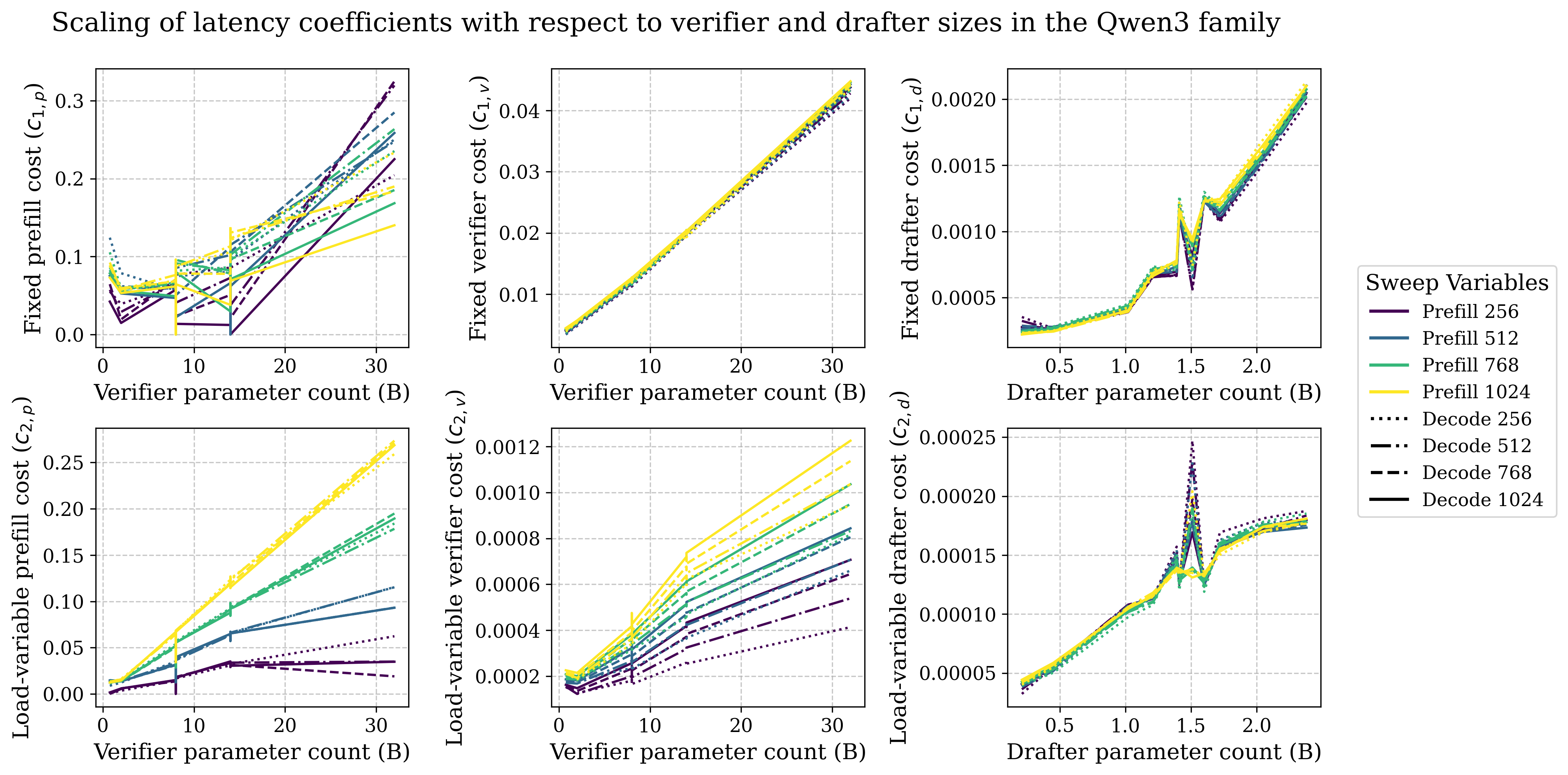}
    \caption{Scaling of SD latency costs with verifier and drafter sizes. A speculation-aware latency model following \cref{eq:intermediate} is fit for each model in the Qwen3 family per prefill and decode combination, but across all \acceptrate{} and \draftlength{}.}
    \label{fig:scaling_params}
\end{figure*}

\paragraph{Fixed-cost coefficients}

The prefill fixed-cost coefficient \conep{} scales roughly linearly with verifier size. It exhibits higher variance and mild non-monotonicity for smaller models, likely because it absorbs batch-size--independent effects such as kernel selection, parallelism choices, and measurement noise. Nevertheless, the dominant trend is linear, especially for medium and large models.

The verifier fixed-cost coefficient \conev{} is nearly perfectly linear in verifier parameter count and is largely invariant across prefill/decode configurations. This is expected because \conev{} captures per-request overheads dominated by model-size--dependent work, such as weight movement and kernel setup, rather than token-length effects.

The drafter fixed-cost coefficient \coned{} similarly scales approximately linearly with drafter parameter count. Deviations are largest when paired with Qwen3-32B, likely due to GPU memory constraints. Increasing drafter depth for Qwen3-8B and Qwen3-14B avoids this bottleneck and confirms the near-linear scaling with drafter size, which is more apparent with Qwen3-32B omitted (\cref{fig:scaling_params_prefill_decode_appendix}).

\paragraph{Load-dependent coefficients}

\begin{figure}[h]
\centering
\begin{minipage}{0.5\textwidth}
    \centering
    \includegraphics[width=\linewidth]{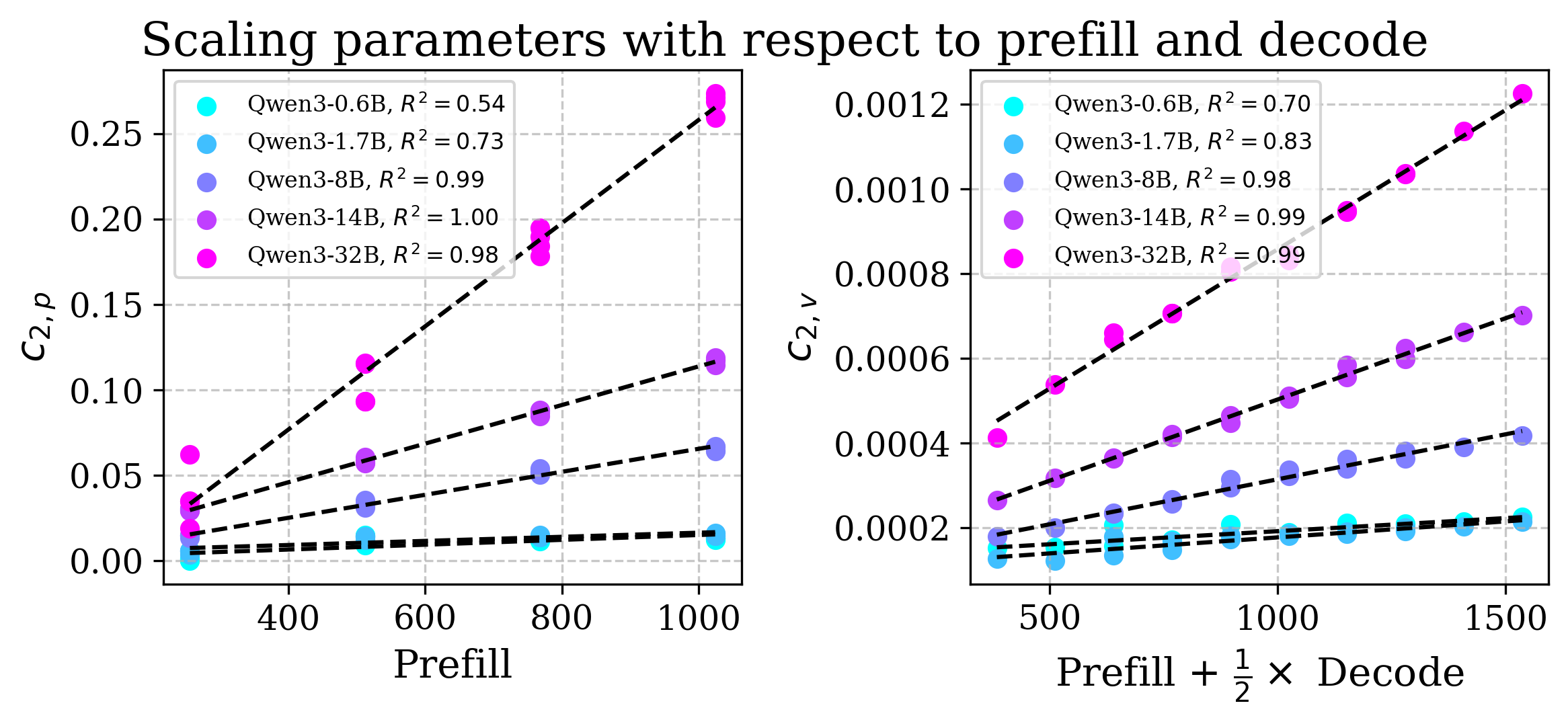}
    \caption{Scaling of SD latency costs with prefill and decode lengths. Same latency models were used as in \cref{fig:scaling_params}.}
    \label{fig:scaling_params_prefill_decode}
\end{minipage}
\begin{minipage}{0.49\textwidth}
    \centering
    \begin{tabular}{|l|ccccc|}
    \hline
        $n$ & 0.6B & 1.7B & 8B & 14B &32B\\
    \hline
    \multicolumn{1}{|c|}{} & \multicolumn{5}{c|}{\underline{\ctwop{}}} \\
        1&0.414&0.644&0.988&0.998&0.978\\
        2&0.407&0.639&0.988&0.998&0.977\\
        4&0.384&0.625&0.987&0.998&0.977\\
    \hline
    \multicolumn{1}{|c|}{} & \multicolumn{5}{c|}{\underline{\ctwov{}}} \\
        1&0.625&0.774&0.971&0.993&0.985\\
        2&0.621&0.771&0.971&0.993&0.985\\
        4&0.610&0.763&0.970&0.993&0.985\\
    \hline
    \end{tabular}
    \caption{Leave-$n$-out analysis for Qwen3.}
    \label{tab:leave_n_out}
\end{minipage}
\end{figure}

The prefill load-dependent coefficient \ctwop{} forms approximately linear trends in verifier size, with slopes increasing for longer prefill lengths. Thus, prefill length mainly rescales the load-dependent cost while preserving linear dependence on model size, suggesting a separable structure further modeled in \cref{fig:scaling_params_prefill_decode}.

The verifier load-dependent coefficient \ctwov{} also scales roughly linearly with verifier size, but depends on both prefill and decode lengths. For fixed prefill length, longer decode lengths increase the slope without changing the linear dependence on model size. Thus, sequence lengths modulate the magnitude of the cost, while model-size scaling remains linear.

The drafter load-dependent coefficient \ctwod{} scales approximately linearly with drafter parameter count. As with \coned{}, the main deviations occur for drafters paired with Qwen3-32B due to memory constraints. Larger drafters paired with smaller verifiers maintain near-linear scaling, indicating that the trend is intrinsic to drafter size (\cref{fig:scaling_params_prefill_decode_appendix}).

\paragraph{Length dependence and generalizability}



For each Qwen model, \ctwop{} is approximately linear in prefill length, with model-specific slopes (\cref{fig:scaling_params_prefill_decode}). Thus, longer prompts primarily increase the magnitude of the load-dependent prefill cost without changing its scaling form.

For \ctwov{}, decode attention cost grows with context length. At decode step $i$, attention scales with $\text{prefill length}+i$, so the average per-token decode cost scales as: $
\text{prefill length}+\frac{1}{2}\text{decode length}$.
This effective token count yields near-linear scaling across model sizes and provides a compact model for verifier load-dependent cost (\cref{fig:scaling_params_prefill_decode}). As shown in \cref{fig:h100}, the same qualitative scaling trends hold on an NVIDIA H100, suggesting that the structure of the coefficient model is robust across GPU architectures even though the fitted values change.

To assess predictive generalization, we perform a leave-$n$-out analysis over the 16 prefill/decode configurations for the key load-dependent terms $c_{2,p}$ and $c_{2,v}$. For each verifier, we hold out all subsets of $n \in \{1,2,4\}$ configurations, fit the coefficient scaling trends on the remaining configurations, and use them to predict the held-out coefficients. The held-out predictive $R^2$ remains close to the full-data fit, especially for larger models (\cref{tab:leave_n_out}). These results suggest that the observed coefficient trends are not merely descriptive fits to measured settings, but retain predictive accuracy for unseen sequence-length configurations.

\subsection{A Mixture of Experts Aware Latency Model} \label{sec:moes}

When fitting MoE models with the same latency model, we find consistently lower latency than predicted for synchronous non-speculative serving. This is expected for sparse experts: at low effective batch size, only a small fraction of experts is activated and loaded, reducing compute and memory traffic relative to dense models. As load increases, more distinct experts are activated, and these savings diminish. Thus, coefficients that are batch-size independent for dense models become load-dependent for MoEs. This gives low latency at light load but rapidly increasing cost under high concurrency, consistent with prior observations on MoE serving difficulty \citep{huang2023towards}.

We capture this effect with an expert-coverage factor $\phi \in [0,1]$, the expected fraction of experts activated in a step $\phi(T)=1-\left(1-\frac{m}{M}\right)^T$,
where $m$ is the number of experts selected per token, $M$ is the total number of experts, and $T$ is the effective routed-token count. This formulation comes from the fact that under independent uniform routing, an expert is never selected with probability $\left(1-\frac{m}{M}\right)^T$. In the non-speculative setting, we approximate $T \approx B = RPS \times L$. We decompose $C_1$ and $C_2$ into low-coverage costs $C_{1,u}, C_{2,u}$ and saturation increments $C_{1,s}, C_{2,s}$. The MoE-aware latency model is

\begin{equation}
\label{eq:moe}
L =
\frac{C_{1,u}+\phi C_{1,s}}
{1-RPS\left(C_{2,u}+\phi C_{2,s}\right)}
\end{equation}

This substantially improves MoE fit, especially at low load where sparse activation is strongest. On the lower-load half of the RPS sweep, $R^2$ improves from $0.902$ to $0.997$ for gpt-oss-20b, from $0.830$ to $0.976$ for Qwen3-30B-A3B, and from $0.906$ to $0.989$ for Qwen3-235B-A22B.

The same effect appears under SD. With short drafts, the verifier processes few tokens per speculative step and activates fewer experts, yielding lower latency than dense scaling predicts. As draft length grows, verifier work and expert coverage increase, making MoE latency closer to dense-model scaling. This interaction explains both the initial deviation and its reduction at larger draft lengths, and may also help explain why MoE speculative speedup underperforms at batch size $1$ \citep{li2024eagle1}.

For SD, we apply the same correction to the verifier terms in Equation~3. Since verifier expert coverage depends on verifier work per speculative step, we use $
T \approx RPS \times L \times k$, yielding $
\phi(T)=1-\left(1-\frac{m}{M}\right)^{RPS \times L \times k}$.

\begin{figure}[h]
    \centering 
    \includegraphics[width=\linewidth]{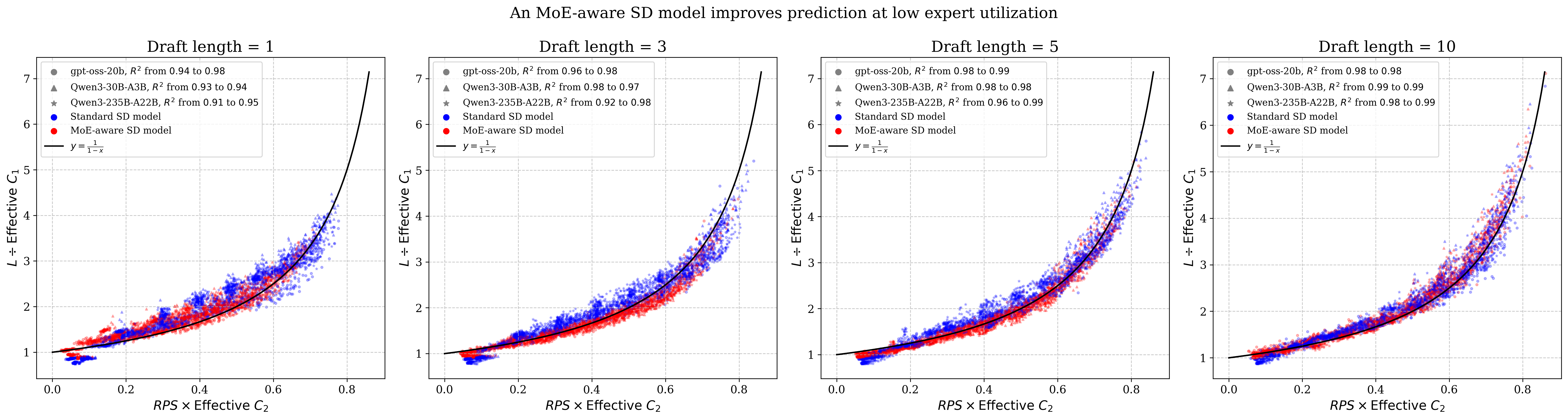}
\caption{
MoE-aware SD latency model improves fit. Compared to the original speculation-aware model, the MoE-aware extension that accounts for load-dependent expert coverage better captures the observed latency behavior, with the largest improvements at low request rates and short draft lengths where sparse expert activation reduces effective verifier cost.
}
    \label{fig:moe_case}
\end{figure}

Substituting this into the verifier coefficients gives
\begin{equation}
\label{eq:moe_sd}
L =
\frac{
c_{1,p}
+\frac{g}{E}\left(c_{1,v,u}+\phi c_{1,v,s}+k c_{1,d}\right)
}{
1-RPS\left(
c_{2,p}
+\frac{g}{E}\left(c_{2,v,u}+\phi c_{2,v,s}+k c_{2,d}\right)
\right)
}
\end{equation}
This again improves the fit over the original model, with the largest gains at low utilization (\cref{fig:moe_case}).

\section{Conclusion and Limitations}
\label{sec:conclusion}

We present an interpretable framework for modeling speculative decoding under continuous-batching LLM serving. By sweeping request rate in a live serving system and using Little's Law to infer effective concurrency, we model latency in the stable pre-saturation regime without treating batch size as an exposed control. SD is incorporated through a prefill/verify/draft cost decomposition, which explains how acceptance rate and draft length reshape fixed and load-dependent latency costs. The model explains a central empirical pattern: SD often provides its largest gains at low load, but speedup erodes as request rate increases unless acceptance rate is sufficiently high. This occurs because speculation typically reduces fixed costs while increasing load-dependent costs, making draft lengths that perform well at batch size one suboptimal under realistic serving load. The fitted coefficients also exhibit structured scaling with model size and sequence length, and the framework extends to MoE models by accounting for sparse expert coverage.

Our study focuses on SD as implemented in current vLLM serving pipelines. Because the model abstracts speculation through the draft-verify cycle, we expect it to extend to other SD variants, though mechanisms such as adaptive drafting or tree verification may introduce additional cost terms. The fitted coefficients are also system-dependent, and different GPUs, parallelism strategies, schedulers, or serving engines may change their numerical values. We therefore view the model as a lightweight profiling framework rather than a universal set of constants: users can measure a small number of configurations on their own stack and assess how SD speedups change with load. Finally, the model targets the stable pre-saturation regime and mean latency. We exclude the saturation boundary, where preemption makes latency unstable. Future work should extend the framework to latency distributions, preemption dynamics, bursty workloads, and broader serving stacks.

Overall, our results suggest that SD should be evaluated and tuned under request-rate load, not only in isolated or fixed-batch settings. The proposed model provides a lightweight framework for assessing when speculation helps, diagnosing when speedup erodes, and reasoning about verifier, drafter, draft length, and architecture choices in deployed serving systems.

\section*{Acknowledgments}

The authors would like to thank Dan Alistarh, Michael Goin, Eldar Kurtic, Dipika Sikka, and Keisuke Kamahori for their helpful discussions and valuable feedback.

This project was supported by an MIT-IBM Watson AI Lab grant.

\clearpage

\clearpage
\bibliographystyle{plainnat}
\bibliography{neurips_2026}
\clearpage


\appendix

\section{Appendix}

\subsection{Derivation of speedup under speculative decoding} \label{sec:speedupderivation}

Let $$\latency{}_D = \frac{C_{1,D}}{1 - \rps{} \times C_{2,D}}$$ be the latency of serving a dense model, along with its associated costs, and let $$\latency{}_{SD} = \frac{C_{1,SD}}{1 - \rps{} \times C_{2,SD}}$$ be the latency of serving a model with speculative decoding. The speedup attainable with speculative decoding is therefore described by:

\begin{equation} \label{eq:speedupfull}
\begin{split}
\text{Speedup}=&\frac{\latency{}_D}{\latency{}_{SD}}\\
&=\frac{C_{1,D}}{1 - \rps{} \times C_{2,D}}\times\frac{1 - \rps{} \times C_{2,SD}}{C_{1,SD}}\\
&=\frac{C_{1,D}}{C_{1,SD}}\times\frac{1 - \rps{} \times C_{2,SD}}{1 - \rps{} \times C_{2,D}}\\
&=\frac{C_{1,D}}{C_{1,SD}}\times\frac{1 - \rps{} \times (C_{2,D} - C_{2,D} - C_{2,SD})}{1 - \rps{} \times C_{2,D}}\\
&=\frac{C_{1,D}}{C_{1,SD}}\times(\frac{\rps{} \times (C_{2,D} - C_{2,SD})}{1 - \rps{} \times C_{2,D}} + 1)\\
&=\frac{C_{1,D}}{C_{1,SD}}\times(\frac{C_{2,D} \times \rps{} \times (C_{2,D} - C_{2,SD})}{C_{2,D} \times (1 - \rps{} \times C_{2,D})} + 1)\\
&=\frac{C_{1,D}}{C_{1,SD}}\times(\frac{C_{2,D} - C_{2,SD}}{C_{2,D}}\times\frac{\rps{} \times C_{2,D}}{1 - \rps{} \times C_{2,D}} + 1)\\
&=\frac{C_{1,D}}{C_{1,SD}}\times(1 + (1 - \frac{C_{2,SD}}{C_{2,D}}\times\frac{\rps{} \times C_{2,D}}{1 - \rps{} \times C_{2,D}})\\
\text{Speedup}&=\frac{1}{C_{1,R}}\times(1 + (1-C_{2,R})\times \frac{r}{1-r})\\
&C_{1,R} = \frac{C_{1,SD}}{C_{1,D}},C_{2,R} = \frac{C_{2,SD}}{C_{2,D}}, r = \rps{} \times C_{2,D}
\end{split}
\end{equation}

\clearpage

\subsection{Scaling speculative decoding without Qwen3-32B}

We fit \cref{eq:intermediate} to all models as before but exclude Qwen3-32B to better visualize the scaling trends of \coned{} and \ctwod{}, as discussed previously (\cref{sec:scaling}). \coned{} and \ctwod{} conform much more closely to a linear relationship with drafter parameter count.

\begin{figure}[h]
    \centering 
    \includegraphics[width=\linewidth]{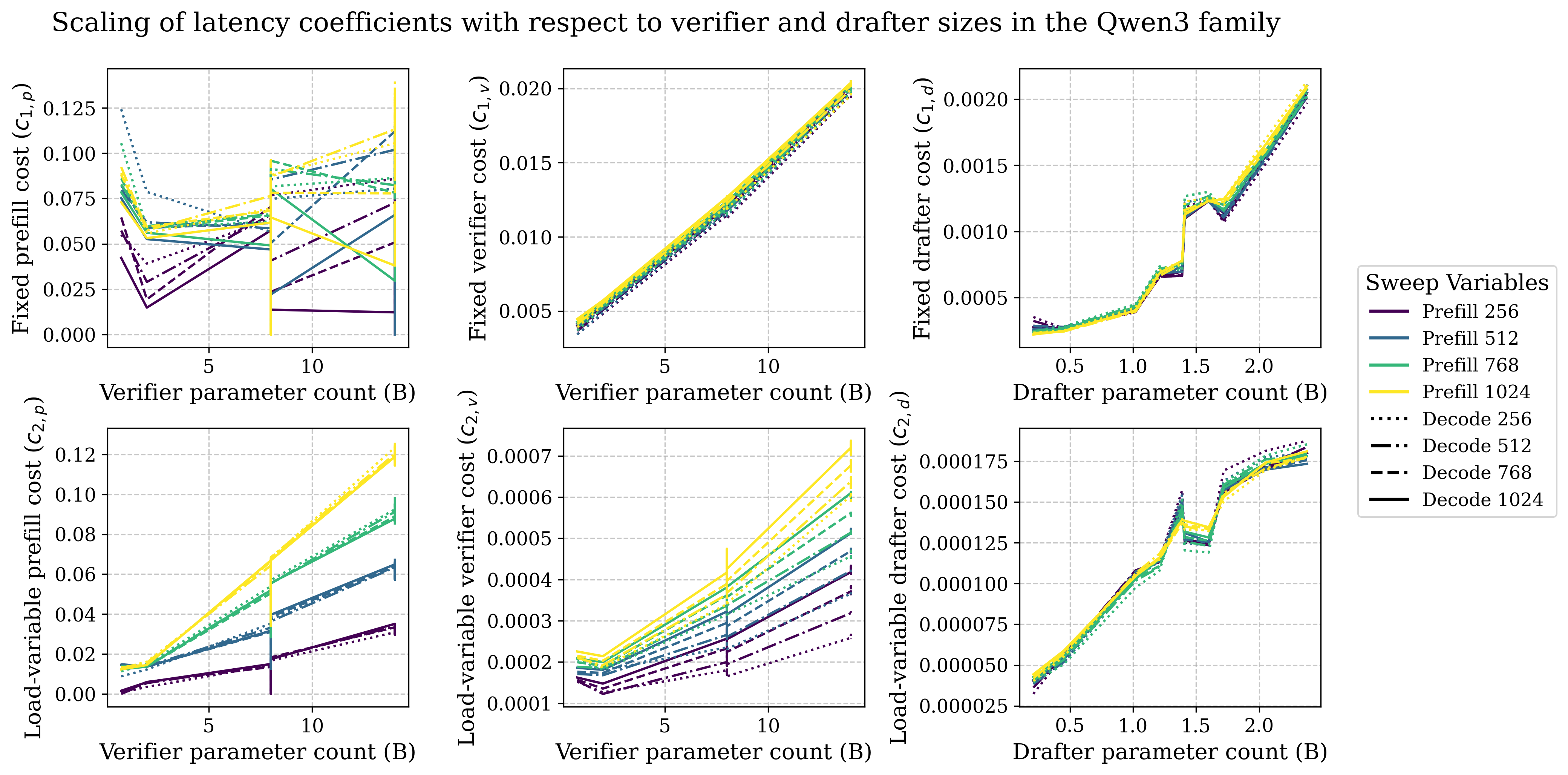}
    \caption{Scaling of speculative decoding latency costs with verifier and drafter sizes.}
    \label{fig:scaling_params_prefill_decode_appendix}
\end{figure}

\subsection{Additional minimum cost ratios}

We show the minimum $C_{1,R}$ and $C_{2,R}$ values attainable for Qwen3 8B for more \acceptrate{} values, extending \cref{fig:speedup_minimums}. 

\begin{figure*}[h]
    \centering 
    \includegraphics[width=\linewidth]{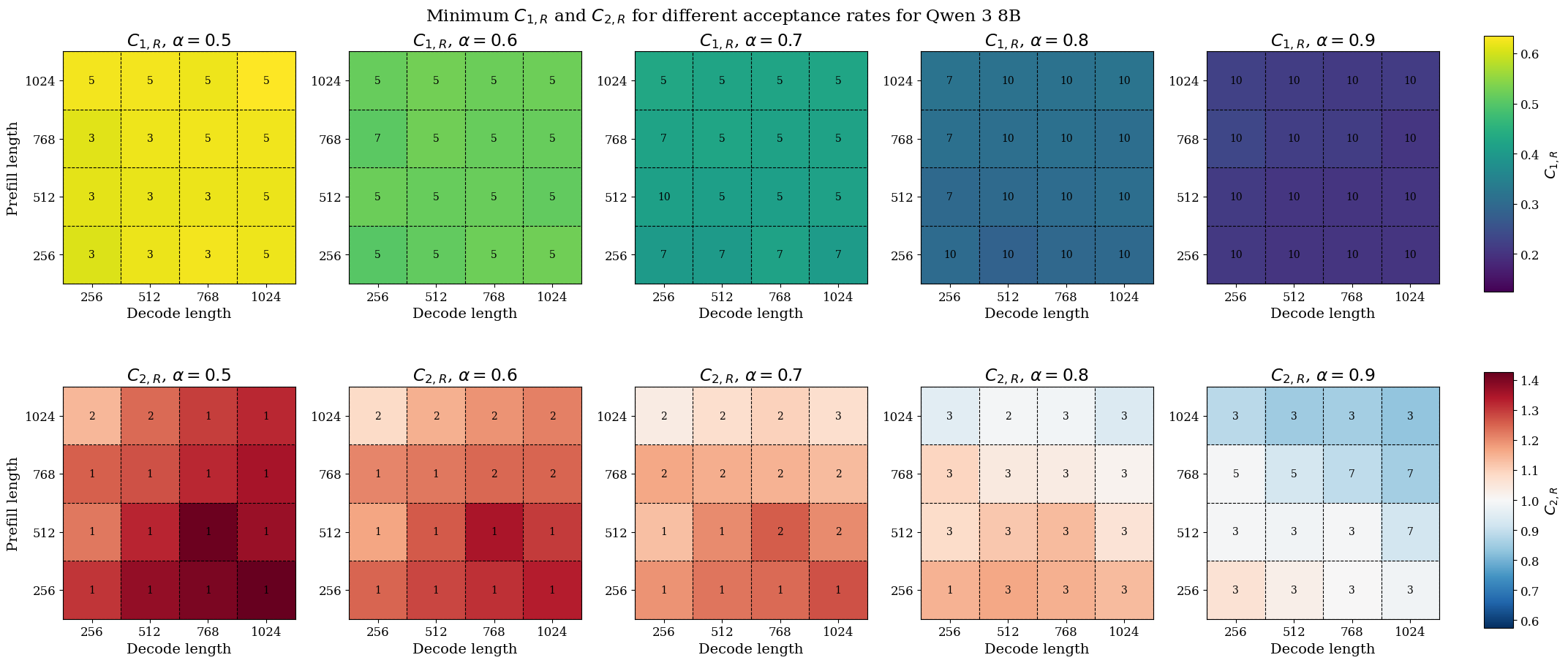}
    \caption{The minimum possible fixed and load-variable cost ratios across prefill and decode configurations. For acceptance rates from $50\%$ to $90\%$, we show the minimum values of $C_{1,R}$ and $C_{2,R}$, and also inscribe which draft length yielded that minimum value.}
    \label{fig:speedup_minimums_full}
\end{figure*}

\clearpage

\subsection{Additional parameter variation for the simple latency model}

We show additional systematic variation of \cone{} and \ctwo{} after each individual speculative decoding configuration is fit by \cref{eq:closed_form_basic} for Qwen3 14B and Qwen3 32B.

\begin{figure*}[h]
    \centering 
    \includegraphics[width=\linewidth]{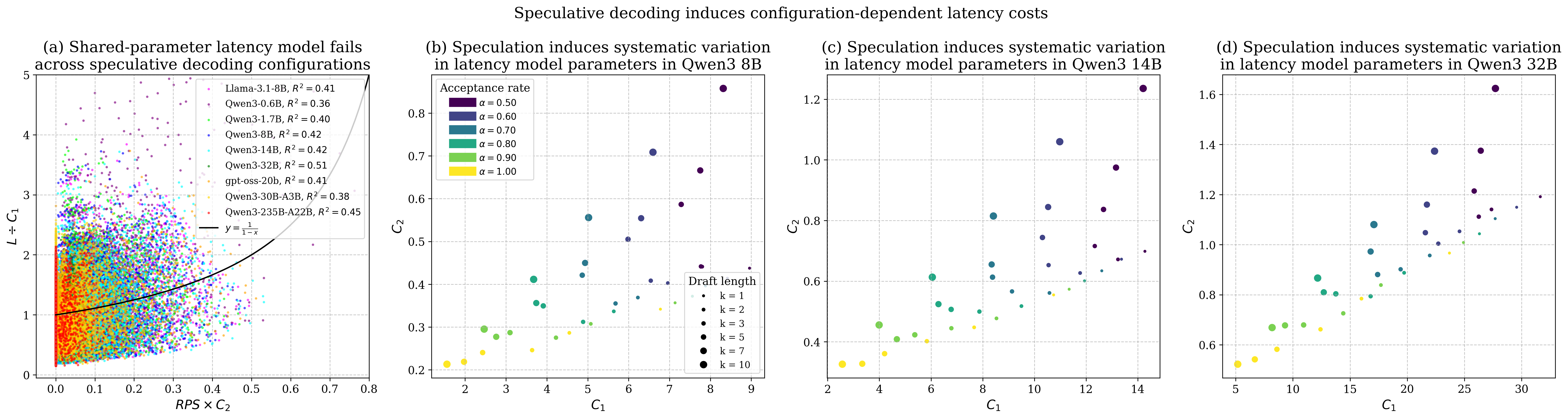}
    \caption{Speculative decoding induces configuration-dependent latency costs. (a) Forcing all acceptance rates and draft lengths to share a single $(\cone{}, \ctwo{})$ yields poor fits, indicating that speculative decoding alters effective service demand, while allowing configuration-specific parameters restores the steady-state latency relationship. Fitted parameters for Qwen3-8B (b), Qwen3-14B (c), and Qwen3-32B (d) vary systematically with acceptance rate \acceptrate{} and draft length \draftlength{}, motivating a speculation-aware cost model. (b-d) were modeled at 1024 prefill and 1024 decode tokens.}
    \label{fig:specdec_basic_latency_full}
\end{figure*}

\clearpage

\subsection{Model robustness for tail latency}
\label{sec:tail_latency}

Our model in \cref{eq:intermediate} is designed for mean latency, since it uses an equilibrium assumption with effective load independent and dependent costs. To assess if this applies to the tail, we refit the model using p95 and p99 end-to-end request latency in place of mean latency and measure the $R^2$ agreement between modeled and observed latency.

\begin{figure*}[h]
    \centering 
    \includegraphics[width=0.6\linewidth]{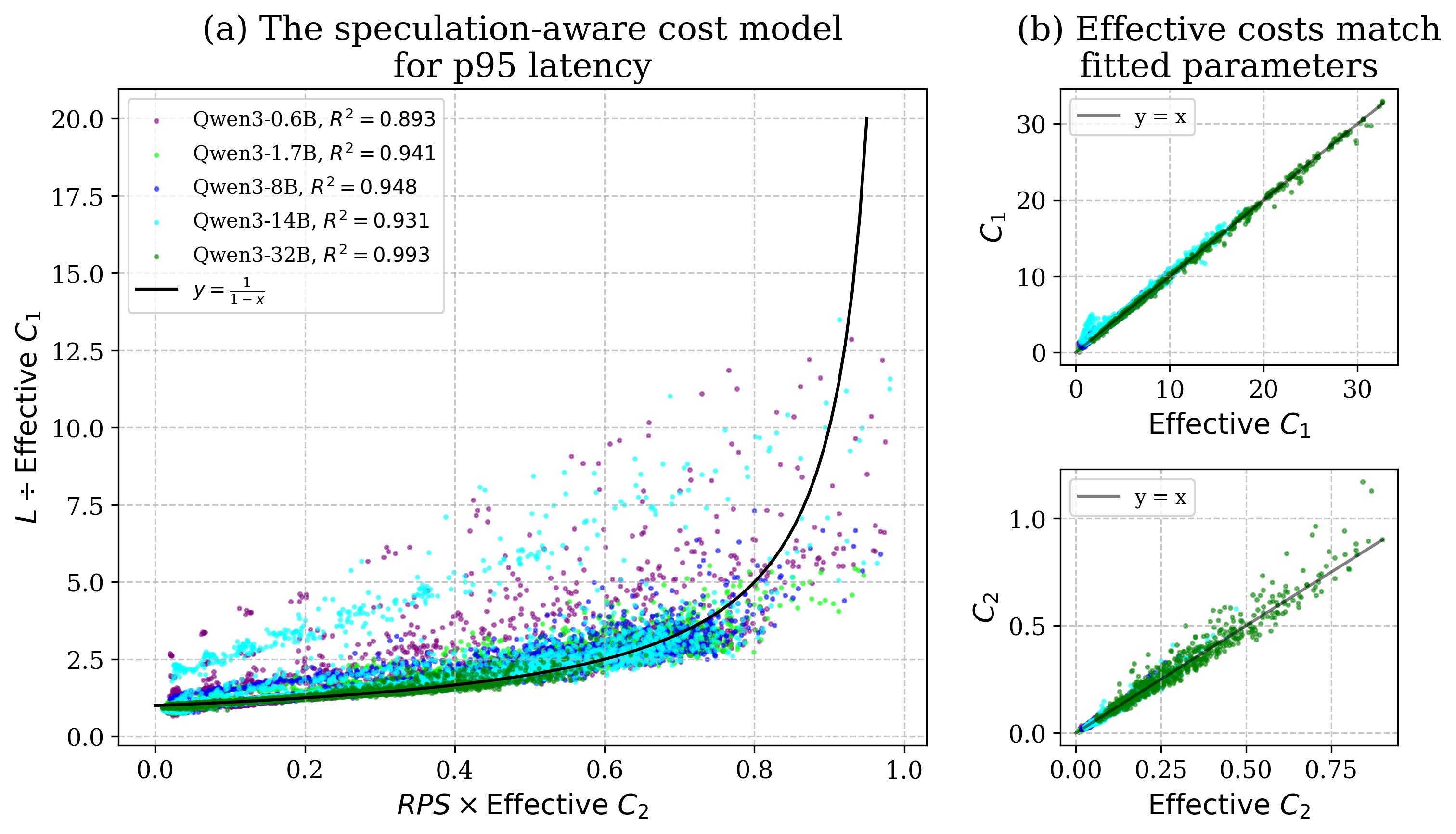}
    \caption{Speculation-aware latency model using p95 latency rather than mean latency.}
    \label{fig:p95}
\end{figure*}

\begin{figure*}[h]
    \centering 
    \includegraphics[width=0.6\linewidth]{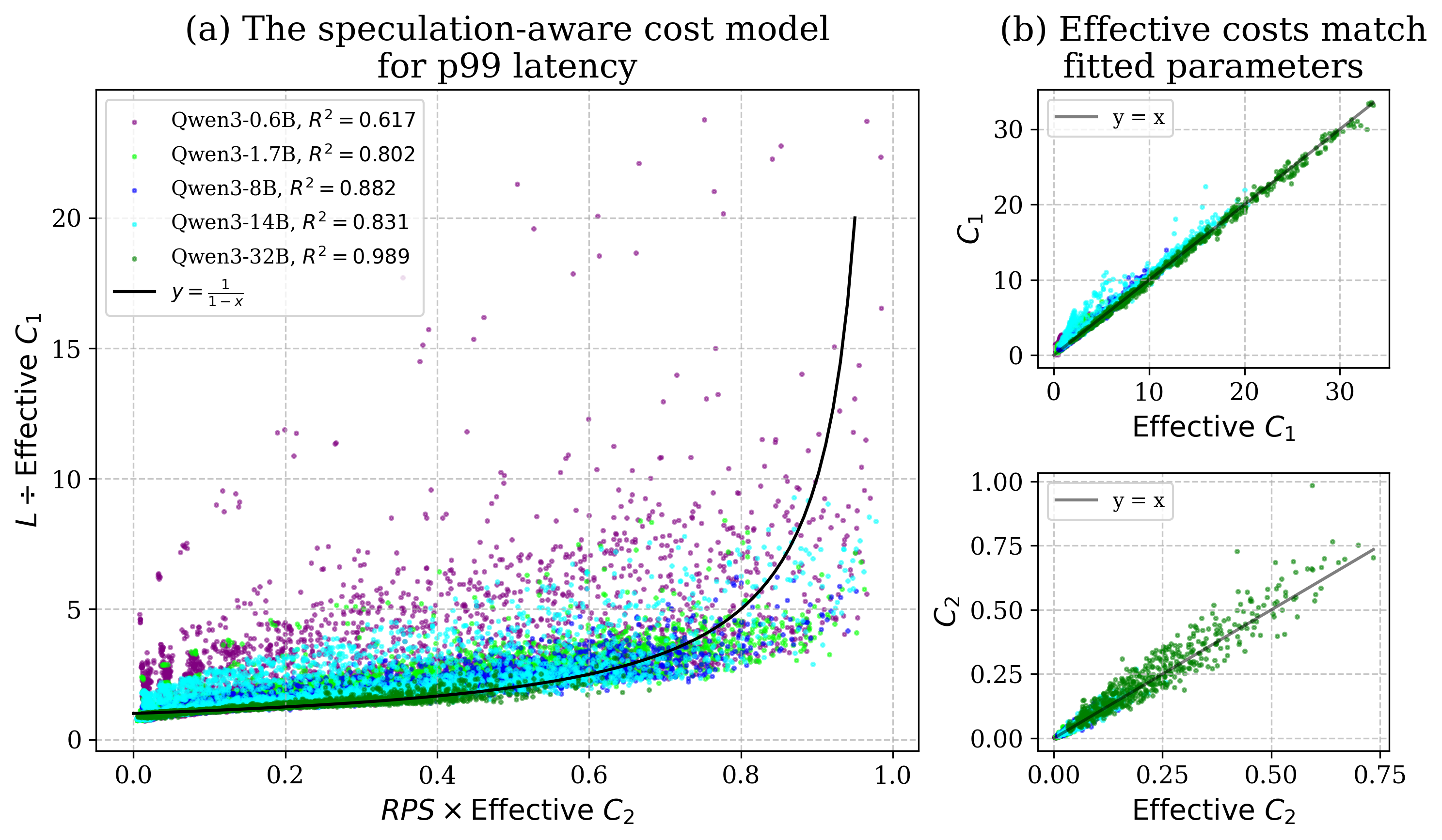}
    \caption{Speculation-aware latency model using p99 latency rather than mean latency.}
    \label{fig:p99}
\end{figure*}

These results suggest that the same steady-state structure remains a good approximation for p95 latency, and for p99 in larger models. We thus view our model as robust to moderate tails. Extending the framework to explicitly model latency distributions is an important direction for future work.

\clearpage

\subsection{Model robustness in different hardware}
\label{sec:h100}

We repeated our experiments on an NVIDIA H100 to test whether the same serving structure holds across GPU architectures. To keep this additional study lightweight, we used a subset of the original prefill/decode settings, about half of the 16 combinations in the main paper. This serves both as a cross-hardware check and as a test of whether the model remains accurate when fit with less data.

We first fit the H100 measurements with \cref{eq:intermediate} and evaluate goodness of fit on the Qwen3 family of models. Even with this subset, the fit remains strong:

\begin{table}[h]
\centering
\begin{tabular}{lccccc}
\toprule
& 0.6B & 1.7B & 8B & 14B & 32B \\
\midrule
$R^2$ & 0.998 & 0.996 & 0.992 & 0.997 & 0.998 \\
\bottomrule
\end{tabular}
\caption{Speculation-aware latency model performance for Qwen3 family on an H100.}
\label{tab:h100}
\end{table}

These results suggest that the same speculation-aware latency model continues to describe the data accurately on an H100.

We also repeated the coefficient scaling experiment on an H100, focusing on the key load-dependent terms $c_{2,p}$ and $c_{2,v}$. The predictive fit $R^2$ is again strong, particularly for the larger models, matching the qualitative trend observed on the A100.

\begin{figure*}[h]
    \centering 
    \includegraphics[width=0.6\linewidth]{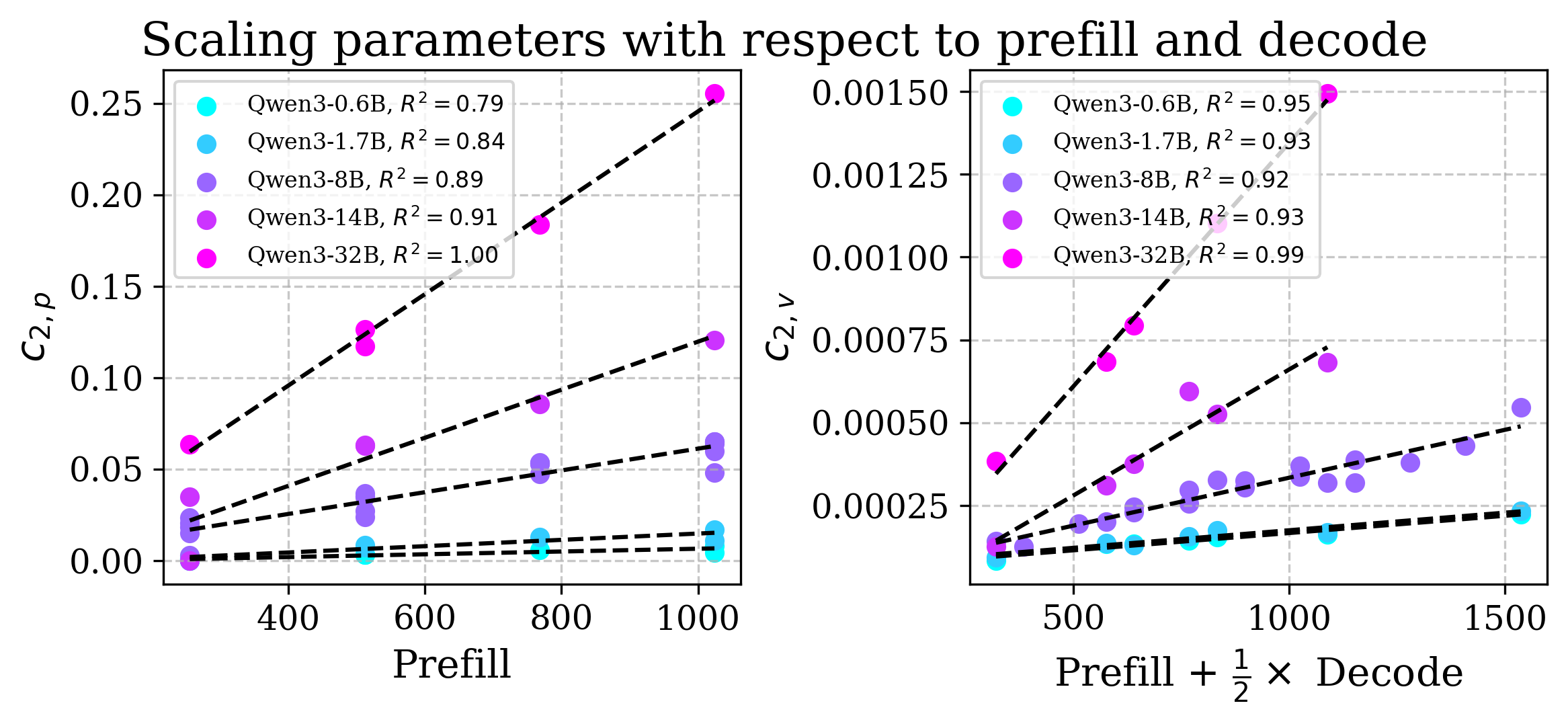}
    \caption{Speculation-aware latency model scaling behavior for Qwen3 family on an H100.}
    \label{fig:h100}
\end{figure*}

Overall, these H100 results support the same conclusion as in the main paper. While changing hardware alters the fitted coefficient values, the underlying load-latency relationship and scaling structure remain the same.



\end{document}